\documentclass[10pt,journal,compsoc]{IEEEtran}

\usepackage[T1]{fontenc}% optional T1 font encoding

\usepackage{graphicx}
\usepackage{amsmath,amssymb}

\usepackage{tabularx}
\usepackage{multirow}
\usepackage{dcolumn}
\usepackage{booktabs}

\usepackage{enumerate}
\usepackage{subfig}

\usepackage{pgfplots}
\pgfplotsset{compat=1.12}
\usetikzlibrary{patterns}

\graphicspath{{./img/}}

\usepackage{nomencl}
\makenomenclature

\usepackage{etoolbox}
\renewcommand\nomgroup[1]{%
  \item[\bfseries
  \ifstrequal{#1}{A}{Acronyms}{%
  \ifstrequal{#1}{S}{Symbols}{%
  \ifstrequal{#1}{U}{Subscripts and Superscripts}{}}}%
]}
\usepackage{tikz}

\usepackage{algpseudocode}
\usepackage{algorithm}

\algnewcommand\algorithmicforeach{\textbf{for each}}
\algdef{S}[FOR]{ForEach}[1]{\algorithmicforeach\ #1\ \algorithmicdo}

\let\oldReturn\Return
\renewcommand{\Return}{\State\oldReturn}

\algnewcommand\algorithmicswitch{\textbf{switch}}
\algnewcommand\algorithmiccase{\textbf{case}}
\algnewcommand\algorithmicassert{\texttt{assert}}
\algnewcommand\Assert[1]{\State \algorithmicassert(#1)}%
\algdef{SE}[SWITCH]{Switch}{EndSwitch}[1]{\algorithmicswitch\ #1\ \algorithmicdo}{\algorithmicend\ \algorithmicswitch}%
\algdef{SE}[CASE]{Case}{EndCase}[1]{\algorithmiccase\ #1}{\algorithmicend\ \algorithmiccase}%
\algtext*{EndSwitch}%
\algtext*{EndCase}%
\usepackage{hyperref}
\hypersetup{
  urlcolor   = blue,
  colorlinks = true,
}

\begin{document}

\title{Deep dual stream residual network with contextual attention for pansharpening of remote sensing images}

\author{
Syeda~Roshana~Ali,
Anis~U.~Rahman,
Muhammad~Shahzad
\thanks{S.R. Ali and A.U. Rahman are with National University of Sciences and Technology (NUST), Islamabad, Pakistan e-mail: \{snaqvi.mscs17seecs,anis.rahman\}@seecs.edu.pk.}% <-this % stops a space
\thanks{M. Shahzad is with Technical University of Munich (TUM), Munich 80333, Germany. email: muhammad.shahzad@tum.de}% <-this % stops a space
\thanks{Corresponding author: A.U. Rahman}
}

% make the title area
\maketitle

%%==================================%%
%% sample for unstructured abstract %%
%%==================================%%

\begin{abstract}
Pansharpening enhances spatial details of high spectral resolution multispectral images using features of high spatial resolution panchromatic image. There are a number of traditional pansharpening approaches but producing an image exhibiting high spectral and spatial fidelity is still an open problem. Recently, deep learning has been used to produce promising pansharpened images; however, most of these approaches apply similar treatment to both multispectral and panchromatic images by using the same network for feature extraction. In this work, we present present a novel dual attention-based two-stream network. It starts with feature extraction using two separate networks for both images, an encoder with attention mechanism to recalibrate the extracted features. This is followed by fusion of the features forming a compact representation fed into an image reconstruction network to produce a pansharpened image. The experimental results on the Pl\'{e}iades dataset using standard quantitative evaluation metrics and visual inspection demonstrates that the proposed approach performs better than other approaches in terms of pansharpened image quality.
\end{abstract}


%%\pacs[JEL Classification]{D8, H51}

%%\pacs[MSC Classification]{35A01, 65L10, 65L12, 65L20, 65L70}

\maketitle

\section{Introduction}
\label{sec:intro}

Multispectral (MS) images have been used for many diverse applications like mineral exploration, crop identification, forest monitoring, land cover mapping, and many more. All these applications take advantage of the high spectral information typically present in MS images. However, the images lack spatial details due to sensor limitations and hardware constraints, and hence, spatial enhancement is required to extract meaningful information. For this purpose, pansharpening is used to fuse high spatial resolution panchromatic (PAN) image with MS image to improve spatial details while preserving spectral details.

Recently, there is interest in deep learning (DL) approaches for pansharpening due to significant performance improvements. A relevant study in~\cite{40} creates an information cube by stacking interpolated MS and PAN images, which is fed to a CNN to learn the mapping between the cube and the reference image for image enhancement. In~\cite{41} a detail injection-based CNN model is used for detailed feature extraction. Similarly, in~\cite{44} a multi-scale and multi-depth CNN-based framework is used for detail extraction. A relevant study in~\cite{45} propose a multi-direction sub-band DL method that trains on patches of high-frequency sub-bands of the PAN image. Other approaches use multi-branch networks for pansharpening, for instance, in~\cite{46,49} such network is used to model high non-linearity and to obtain high-level features for fusion tasks. Furthermore, recent deep attention approaches incorporate contextual knowledge that has resulted in improvements for different vision tasks including pansharpening. In~\cite{66}, a spatial-channel attention pansharpening approach learns mapping between image pairs based on differential information. In~\cite{67} a residual CNN uses a channel attention mechanism treating each channel independently. 

% CNN
% Guo et al.~\cite{42} propose a CNN-based model with dilated multilevel blocks to extract features more efficiently. 
% DCNN
% Eghbalian et al.~\cite{43} presented a DCNN model to estimate the detailed injection as in MRA techniques in addition to the proposed special loss function that focuses on learning the spectral characteristics. 
% CNN

% CNN
% Chen et al.~\cite{47} also presented an image-to-image CNN model for feature extraction.

% between a multi architecture a multichannel DCNN model for pansharpening that works as channel level enhancement and helps to deal with the problem of spectral mismatch between the MS and the PAN image. 

% Similarly, in~\cite{50} a two-branch multi-level CNN model fuses MS and PAN images to achieve high nonlinearities and obtain high-level features for fusion tasks. 

% Gaetano et al.~\cite{48} presented a simple two-branch CNN model for the fusion of MS and PAN images. 

% Liu et al.~\cite{49} also proposed a two-stream CNN based fusion framework in which one stream consists of two branches to extract features from both the input images while the second stream is used to combine the extracted features and finally the enhanced image is obtained by fusing the combined features. 

% such as classification~\cite{53}, image dehazing~\cite{56}, recognition~\cite{57}, super resolution~\cite{60} and enhancement~\cite{65} etc. 

% Numerous researchers have used attention mechanisms for pansharpening as well. 

% In~\cite{68} Yang et al. proposed a novel spatial attention-based approach for fusion of HS and MS images. 

Even though the DL approaches significantly improve results, but many of these approaches treat all extracted features as the same by learning the same set of model parameters. Recall that the MS and PAN images are highly correlated, but still they have different information that is equally important. Here, attention can effectively overcome this problem but most of the pansharpening approaches apply attention after combining both MS and PAN images or using them in isolation. In this study, we present a novel two-stream approach based on dual attention for MS pansharpening. That is, we use two separate neural networks with different layers having separate kernels for comprehensive feature extraction, enabling the model to flexibly deal with various levels of detail. Furthermore, inspired by the efficacy of attention, we use lightweight pixel- and channel-level attention blocks to reweight and recalibrate respective features. Lastly, we introduce multiple long and short skip connections and residual learning to minimize spectral and spatial loss in the pansharpened images. 

% Consequently, the proposed approach enhances the quality of pansharped images without increasing the overall computational cost. Lastly, extensive experiments are carried out on the Pl\'{e}iades dataset consisting of images belonging to two different locations. The computed value of quality metrics indicates that the overall proposed approach performed well than the various presented approaches.

% The main contributions are as follows:
% \begin{itemize}
% \item Propose a novel end-to-end dual attention-based two-stream (DATS) approach for pansharpening of multispectral images with feature extraction and fusion carried out in the feature domain.
% \item Extract features are extracted first from both the input images (PAN and MS) using two different networks having different layers and kernels.
% \item To further enhance the extracted distinct features, the channel level, and pixel-level attention mechanism is employed.
% \item The concept of multiple long- and short-skip connections and the residual learning is also incorporated that enables the network to bypass the low-frequency details with the help of shortcut connections and maintain the balance among the distinct information.
% \item To further optimize the proposed approach, we used $\ell_1$ as a loss function as an alternative to the generally used $\ell_2$ loss function. This substitution considerably improves the fusion quality.
% \end{itemize}

The rest of the paper is organized as follow: Section~\ref{sec:proposed} formulates the pansharpening problem and elaborates the proposed architecture. Section~\ref{sec:results} describes the experimental evaluation followed by discussion of quantitative and visual results. Section~\ref{sec:conclusions} concludes the work.

% \newpage

\section{Problem Formulation and Proposed Framework}
\label{sec:proposed}

% Inspired by the use of CNN in many other image processing domains and its powerful capability of representing an image into its hierarchal feature and then reconstructing image back from those features flawlessly, 

\begin{figure*}[!ht]
\centering
\includegraphics[width=1.0\linewidth]{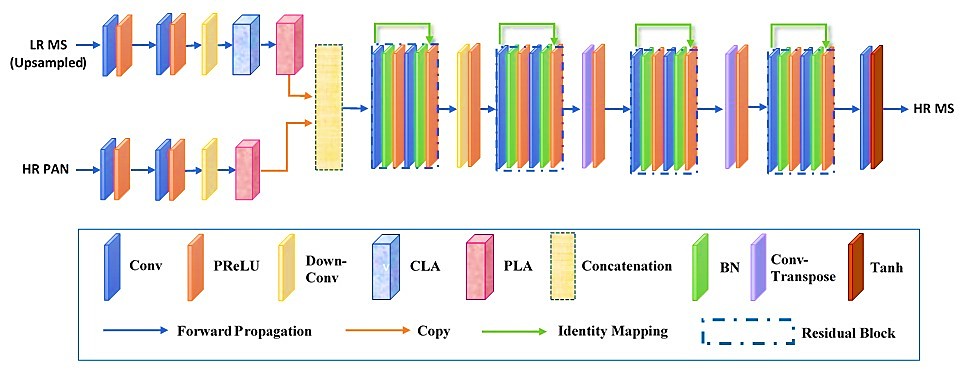}
\caption{Detailed architecture of proposed dual attention-based two-stream pansharpening approach.}
\label{fig4}
\end{figure*}

We propose a pansharpening method that deals separately with both PAN and MS image for feature extraction with help of attention mechanism. The features are fused followed by reconstruction of a high-resolution MS image. Fig.~\ref{fig4} illustrates the complete architecture of the proposed pansharpening approach.

% The architecture of the proposed model is depicted in Fig.~\ref{fig1}.

% \begin{figure}[!ht]
% \centering
% \includegraphics[width=0.9\linewidth]{img/1a.jpg}
% \caption{Conceptual framework of the proposed dual attention-based pansharpening approach.}
% \label{fig1}
% \end{figure}

\subsection{Feature Extraction}
Similar to any vision task, pansharpening can use feature extraction methods from hand-crafted features to sparse encoding. However, the former are categorized as lossy transformations making it unsuitable for image reconstruction, while the latter requires a high-resolution image for dictionary creation and training. In contrast CNNs are capable of extracting image features, as well as, combining them in the feature domain.

% are proposed to extract these features from corresponding images. First are hand-crafted features for instance SIFT, LBP, and HOG, but they do not fulfill the requirement of reconstructing the original image from the obtained features. In other words, these techniques can be categorized as lossy transformations. Another most used feature crafting technique is sparse encoding, which utilizes a dictionary created during training to represent an image. There are several sparse encodings-based pansharpening approaches proposed but the fusion is still carried out in the image domain. Because fusing features in the sparse domain is still uncertain. Also, this technique requires a high-resolution image during training for dictionary creation, which is not available in most cases. So, this technique will not work well where a high-resolution image is not available for training. In CNN has the power to extract features from the image and then can also combine them in the feature domain.

\subsection{Attention Mechanism}
% Many studies have shown promising results by using attention mechanisms in the construction of neural networks. Recently it is being employed for a leading image restoration task i.e. super resolution~\cite{60}. Many researchers have used it for pansharpening as well~\cite{66}. 

% Inspired by their work we have also incorporated attention in our model but separately for each image to deal with the distinct information contained in them. 

The resulting feature maps are processed using an attention mechanism to further enhance them. Recall that PAN and MS images contain different information, and hence, we design two types of attention mechanisms: pixel-level attention for PAN image and channel-level attention for MS image. Consequently, this assigns distinct weights to features at different levels before passing them to the feature fusion network.

\begin{figure}[!ht]
\centering
\includegraphics[width=1.0\linewidth]{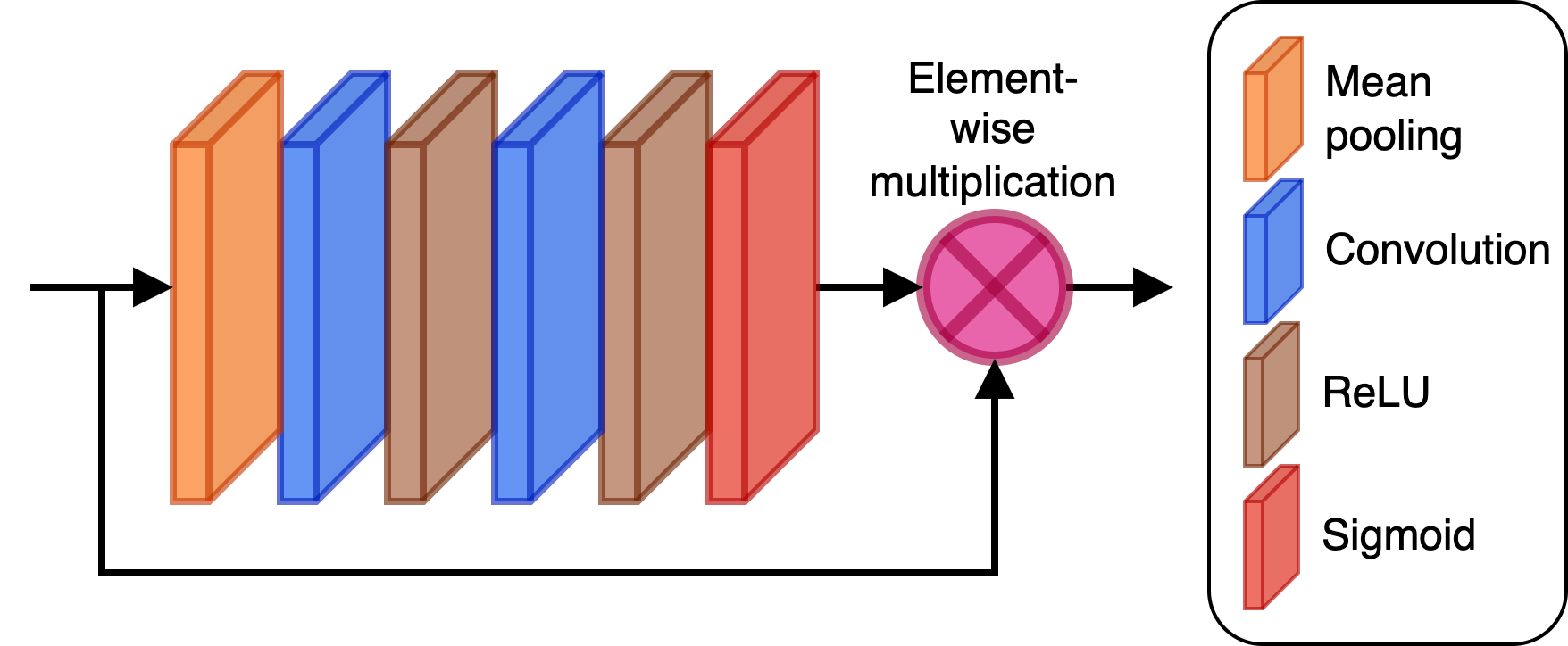}
\caption{Layout of channel-level attention module.}
\label{fig2}
\end{figure}

\textbf{Channel-level attention.} recovers lost spatial information using a global mean pooling function (Fig.~\ref{fig2}), formulated as:
\begin{align}
 \breve{G}_c= F_c ( \breve{E}_c)=  \frac{1}{h\times w}\sum_{i=1}^h \sum_{j=1}^w{I_c (i,j)}
\end{align}
where $I_c (i,j)$ is the pixel at $(i, j)$ in channel $c$, and $F_c$ is the mean pooling function. The transformed feature map---from $(h \times w \times c)$ to $(1 \times 1 \times c)$---is fed to a dual-layer convolutional model with ReLU activation after each convolutional layer and sigmoid at the end, formulated as:
\begin{align}
W_c=\sigma (conv(\Upsilon(conv(\breve{G}_c))))
\end{align}
where $\Upsilon$ and $\sigma$ demotes ReLU and sigmoid activations, respectively. The last step is element-wise multiplication of $\breve{G} (I_M)$ and $W_c$ depicted as:
\begin{align}
 \breve{E}_{CLA}(I_M) =  \breve{E}(I_M)\otimes W_c
\end{align}

\begin{figure}[!ht]
\centering
\includegraphics[width=1.0\linewidth]{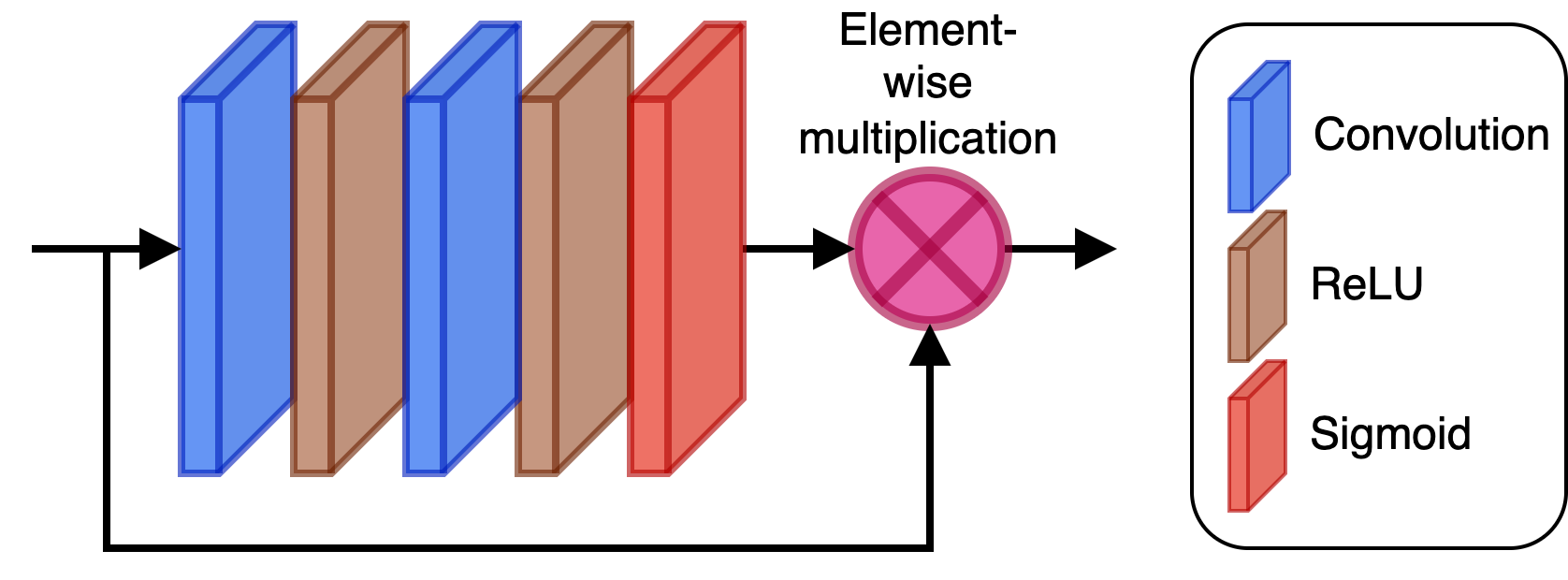}
\caption{Layout of pixel-level attention module.}
\label{fig3}
\end{figure}

\textbf{Pixel-level attention.} Similarly, information varies among different image pixels. We explicitly capture this information using a pixel-level attention (PLA) module. Here, the feature maps obtained from the PAN image feature extractor are fed to a dual-layer convolutional model with ReLU activation after each convolutional layer and sigmoid at the end as illustrated in Fig.~\ref{fig3}. The model computes weights for distinct pixels as:
\begin{align}
W_p=\sigma (conv(\Upsilon(conv(\breve{E}(I_P)))))
\end{align}
The last step of this module is element-wise multiplication of $\breve{E}(I_P)$ and $W_P$ depicted as:
\begin{align}
 \breve{E}_{PLA}(I_P) =  \breve{E}(I_P)\otimes W_P
\end{align}

Notice that the PLA module is also applied to an MS image after CLA module to further enhance channel-level features. 

\subsection{Fusion Module}
Once the feature maps are enhanced, the next step is their fusion. Generally, some pooling technique like mean or max pooling is used to fuse these feature maps, but this leads to information loss, contrary to the pansharpening task where every bit of information needs to be retained. To avoid this, we use an effective alternative for pooling that is concatenation, formulated as:
\begin{align}
 \breve{E}(I_p, I_M) =  \breve{E}_{CLA}(I_M) \oplus \breve{E}_{PLA}(I_P) 
\end{align}
where $\oplus$ denotes the concatenation, and $ \breve{E}(I_p, I_M)$ is the fused feature map. This fused feature map acts as an input to the fusion network, which creates a more precise encoded representation of the concatenated features using three convolutional layers. The result is a tensor sized $[\frac{w}{4}, \frac{h}{4}, 256]$, a compact encoding of spectral and spatial details of the MS and PAN images. This tensor is the input for the final reconstruction network used to reconstruct the pansharped image.

\subsection{Reconstruction Network}
The feature maps contain only one-fourth of the input in terms of height and width. Even though a simple linear interpolation technique can be used to match the spatial resolution requirement but a learnable approach would perform much better~\cite{72}. Here, we adopt a learnable approach for upsampling in terms of spatial resolution, a reconstruction network comprising transposed convolutional layers. 

\subsection{Skip Connections}
% Generally, high-level features are used for recovery of information while they often fail to provide the texture details because they only contain abstract details of the input~\cite{73}. 

% Recently residual learning approach~\cite{74} is being effectively utilized in many computer vision tasks~\cite{76}. 

To obtain a high quality pansharpened image, we include information from all levels of representation, but this increases the model's computational complexity. To alleviate this, we use skip connection to copy feature maps from the encoder directly to the decoder after each upsampling phase, concatenating them to equivalent feature maps to infuse any lost details during downsampling. Here, we use a residual network (ResNet) to aid the flow of information from low-level layers to high-level layers. The network uses residual CNN units instead of simple CNN units, expressed as:
\begin{align}
x_n = \hat {I} (r_n) + \Omega (r_n, \omega_n); r_{n+1} =\phi (x_n)
\end{align}
where $r_n$ and $r_{n+1}$ are input and output of ${n}^{th}$ residual CNN unit, $\Omega$ is residual function, $\hat{I}(r_n)$ is identity mapping, and $\phi (x_n)$ is activation. 

% All weights in the model are learned adaptive learned that works much better than the one specified directly. 

\subsection{Loss Function}
The integral part of training a neural network is the loss function that computes the model error by evaluating a set parameter $(\vartheta)$. In the case of pansharpening the loss function also significantly affects the quality of pansharped image. Often $\ell_2$ norm is used as loss function for image-related tasks; however, this causes blurriness in the restored image. Instead, loss function $\ell_1$ is used for better training of image restoration models~\cite{77}. Likewise, we formulate the loss function $\ell_1$ for optimal training of the proposed network as:
\begin{align}
\ell_1 (\vartheta) = \frac{1}{n}\sum_{i=1}^n \mid  \breve{E}(I_P^i , I_M^i; \vartheta)- X^i\mid_1
\end{align}
where $I_P$ is PAN image, $I_M$ is low-resolution MS image (LrMS) and $X$ is high-resolution MS image (HrMS), and $n$ is the training batch size.

% The value of these weight parameters is optimized by minimizing the model error. 

\section{Results}
\label{sec:results}

% This section of the manuscript summarizes the experimental work done for the performance evaluation of the proposed approach. Different quality measures are computed to assess the performance of the proposed model are also discussed. Visual and quantitative results are compared with a recently proposed approach.

\subsection{Experimental Dataset}
Experiments are carried out on the Pl\'{e}iades dataset. Images of two different locations: Toulouse and Strasbourg (France), are used for training and testing. These images contain trees, towns, roads, vegetation areas, water areas, and other kinds of land cover. The choice enables the evaluation of robustness of the proposed approach. 

% Table~\ref{tab1} lists the general details of the dataset.

% \begin{table}[!ht]%
% % \footnotesize
% \centering
% \caption {Attributes of the Pl\'{e}iades dataset used for training and testing. \label{tab1}}
% \begin{tabular*}{1.0\linewidth}{@{\extracolsep\fill}lccccccccccclD{.}{.}{3}l@{\extracolsep\fill}}
% \toprule
% Year & 2011 \\ \midrule
% Mission type & Earth observation \\ \midrule
% Resolution   & Panchromatic 0.8m \\ 
% ~ & Multispectral 3.2m \\ \midrule
% Bands (N) & P: 470--830nm \\ 
% ~ & Red: 590--710nm\\ 
% ~ & Green: 500--620nm \\ 
% ~ & Blue: 430--550nm \\ 
% ~ & Near-infrared: 740--940nm \\ \midrule
% Dimension    & 1024 $\times$ 1024 \\ \midrule
% Swath width & 20km \\
% \bottomrule
% \end{tabular*}
% \end{table}

\subsection{Evaluation Metrics}
Recall that pansharpening fuses two images with different resolutions to obtain a single high-resolution image but due to the different dimensions, it introduces spectral and spatial distortions in the fused image. The robustness of the pansharpening method is evaluated by analyzing the fused image against the reference high-resolution MS image. Considering $\hat{H}$ is the fused image and $H$ is the reference image, both belonging to $\mathbb{R}^{M\times N}$, where $i$ and $j$ correspond to image row and column indices, and $\mu$ and $\sigma$ denote the mean and standard deviation of the images. The quality metrics used for evaluation are as follows:

\begin{enumerate}
\item \textbf{Erreur relative globale adimensionnelle de synthèse (ERGAS).}~\cite{64} measures image quality by taking normalized mean error of every band of fused image, given as:
\begin{align}
ERGAS = 100\frac{dk}{dl}\sqrt{\frac{1}{n}\sum_{i=1}^n\left( \frac{RMSE_i}{mean_i}\right)^2}
\end{align}
The ideal value is 0 whereas a higher value indicates distortions in the fused image.

\item\textbf{Spectral angular mapper (SAM).}~\cite{85} calculates the spectral angle between referenced image and fused image pixels. The metric measures the preserved spectral details by taking the mean of all the values, given as:
\begin{align}
SAM=\arccos\left(\frac{H_j,\hat{H}_j)}{\parallel H_j\parallel,\parallel\hat{H}_j\parallel}\right)
\end{align}
Here, the output is either in radians or degrees with an ideal value of 0.
 
\item\textbf{Universal image quality index (UIQI).}~\cite{60} computes the transformation of known data from the original image into a pansharpened image, given as:
\begin{align}
UIQI=\frac{4\sigma_{\hat{H}H}(\mu_{H}+ \mu_{\hat{H}})}{(\mu_{H}^2+ \mu_{\hat{H}}^2)(\mu_{H}^2+ \mu_{\hat{H}}^2)}
\end{align}
The values range between -1 and 1, an ideal value of 1 for same images.

\item\textbf{Spatial correlation coeficient (SCC).}~\cite{88} assesses the spatial quality of pansharpened images. Here, the spatial details of pansharpened and reference images are extracted and then compared by computing a correlation coefficient between the extracted details. A Laplacian filter is computed band-by-band for detail extraction and correlation. The ideal value is 1 indicating spatial correlation between the two images.

\item\textbf{Structural similarity index measure (SSIM).}~\cite{68} is a similarity criterion between fused and reference images in terms of structure, contrast, and luminance, given as:
\begin{align}
SSIM=\frac{(2\mu_{H}\mu_{\hat H} + C_{1})(2\sigma_{H \hat H}+C_{2})}{({\mu_H}^2 + {\mu_{\hat H}}^2 + C_1)({\mu_H}^2 +{\mu_{\hat H}}^2 +C_2)}
\end{align}
\end{enumerate}
The range of value is -1 to 1 with an ideal value of 1 representing similar images.

\subsection{Methods for comparison}
For comparison, we use several conventional methods including IHS~\cite{7}, PCA~\cite{8}, BDSD~\cite{80}, PRACS~\cite{82}, indusion~\cite{84}, GLP~\cite{15}, CBD~\cite{85}, HPM~\cite{16} and HPM-PP~\cite{86}. From the DL domain, we use CAE~\cite{87} and TFNet~\cite{49}. 
% The implementations are done as mentioned by the authors with tests performed under similar conditions.

% BT 81 1987
% PCA 8 1989
% GS 12*
% HPF 83 1991

\subsection{Implementation details}
There are 50,800 training image patches (LrMS, HR PAN, upsampled LrMS, and HrMS) of size 128$\times$128. The implementation is done using PyTorch on an NVIDIA GeForce GTX 980 GPU. The batch size is set to 32, adam optimizer is used for minimizing the loss with a 0.5 and 0.0001 momenta and learning rate, respectively. The model training takes around 12 hours.

% The larger sizes like 256/512 of an image can also be supported by our model, but it requires more memory so to avoid complexities optimal size is selected. 

\subsection{Experimental analysis}
% In this section, we will compare our model with some conventional and recently proposed pansharpening approaches with the help of different experiments and report the quality metrics computed against each experiment along with visual results. 

We conduct two experiments to evaluate the performance of the proposed approach under different conditions. In first experiment, we consider the most favorable case of evaluation, where training and test images are of the same location taken by the same sensor. We use images of the Pl\'{e}iades dataset representing urban areas of Toulouse and Strasbourg (France). Table~\ref{tab2} and~\ref{tab3} presents different quality metrics for different approaches including the proposed approach. Here, the low values of the first two metrics: ERGAS and SAM, indicate less distortion and spectral fidelity in the proposed approach, whereas the high value of UIQI and SSIM demonstrates that the learned features are similar to the reference image. Moreover, the ideal value achieved for SCC indicates that the pansharpened image from the proposed approach is spatially correlated to the reference image. TFNet also performs well producing better results compared to all conventional approaches. Notice that indusion method suffers from high distortion, meaning the technique lacks in its ability of feature representation. 

% , along with their ideal values for two experiments carried out under favorable case. Optimal values obtained are presented in bold.

\begin{table}[!ht]%
\footnotesize
\centering
\caption{List of quality metrics for a favorable case on Toulouse-Pl\'{e}iades dataset. \label{tab2}}
\begin{tabular*}{1.0\linewidth}{@{\extracolsep\fill}lccccccccccclD{.}{.}{3}l@{\extracolsep\fill}}
\toprule
Methods & ERGAS & SAM & UIQI & SCC & SSIM \\
\midrule
% Reference & \textbf{0}      & \textbf{0}      & \textbf{1}      & \textbf{1}      & \textbf{1}      \\ 
IHS                & 3.005          & 0.111          & 0.983          & 0.665          & 0.848          \\ 
PCA                & 2.882          & 0.105          & 0.984          & 0.666          & 0.859          \\ 
BDSD               & 2.868          & 0.094          & 0.990          & 0.660          & 0.885          \\ 
BT                 & 3.0526          & 0.1137          & 0.9834          & 0.6665          & 0.840          \\ 
GS                 & 2.9697          & 0.1081          & 0.9827          & 0.6655          & 0.855          \\ 
PRACS              & 2.419          & 0.080          & 0.992          & 0.665          & 0.882          \\ 
HPF                & 3.7355          & 0.1367          & 0.9830          & 0.3475          & 0.713          \\ 
Indusion           & 5.945          & 0.218          & 0.933          & 0.411          & 0.445          \\ 
GLP            & 2.706          & 0.093          & 0.991          & 0.664          & 0.874          \\ 
CBD        & 2.723          & 0.093          & 0.990          & 0.664          & 0.872          \\ 
HPM        & 2.723          & 0.093          & 0.990          & 0.667          & 0.874          \\ 
HPM-PP     & 3.027          & 0.106          & 0.988          & 0.662          & 0.857          \\
CAE                & 2.856          & 0.092          & 0.985          & 0.626          & 0.868          \\ 
TFNet              & 0.657          & 0.024          & 0.999          & 0.872          & 0.986          \\ 
DATS               & \textbf{0.640} & \textbf{0.024} & \textbf{1.000} & \textbf{0.878} & \textbf{0.988} \\
\bottomrule
\end{tabular*}
\end{table}

\begin{table}[!ht]%
\footnotesize
\centering
\caption{List of quality metrics for a favorable case on Strasbourg-Pl\'{e}iades dataset. \label{tab3}}
\begin{tabular*}{1.0\linewidth}{@{\extracolsep\fill}lccccccccccclD{.}{.}{3}l@{\extracolsep\fill}}
\toprule
Methods & ERGAS & SAM & UIQI & SCC & SSIM \\
\midrule
% Reference        & \textbf{0}      & \textbf{0}      & \textbf{1}      & \textbf{1}      & \textbf{1}      \\
IHS              & 1.639          & 0.063          & 0.997          & 0.452          & 0.754          \\ 
PCA              & 1.431          & 0.054          & 0.997          & 0.454          & 0.905          \\
BDSD             & 1.850          & 0.062          & 0.996          & 0.433          & 0.891          \\ 
BT               & 1.7582          & 0.0669          & 0.9963          & 0.4534          & 0.7356          \\ 
GS               & 1.4461          & 0.0509          & 0.9970          & 0.4543          & 0.8977          \\ 
PRACS            & 1.414          & 0.048          & 0.997          & 0.456          & 0.913          \\ 
HPF              & 2.2086          & 0.0747          & 0.9940          & 0.2167          & 0.7654          \\ 
Indusion         & 4.041          & 0.1352          & 0.971          & 0.260          & 0.438          \\ 
GLP          & 1.708          & 0.0593          & 0.997          & 0.446          & 0.834          \\ 
CBD      & 1.707          & 0.0580          & 0.997          & 0.446          & 0.876          \\ 
HPM      & 1.703          & 0.0593          & 0.997          & 0.445          & 0.834          \\ 
HPM-PP   & 1.875          & 0.0652          & 0.996          & 0.445          & 0.823          \\
CAE              & 2.540          & 0.0525          & 0.989          & 0.438          & 0.874          \\ 
TFNet            & 0.708          & 0.0231          & 0.999          & 0.614          & 0.983          \\ 
DATS             & \textbf{0.701} & \textbf{0.023} & \textbf{0.999} & \textbf{0.616} & \textbf{0.984} \\
\bottomrule
\end{tabular*}
\end{table}

\begin{figure*}[!htb]
\centering
\subfloat{\includegraphics[width=0.21\textwidth]{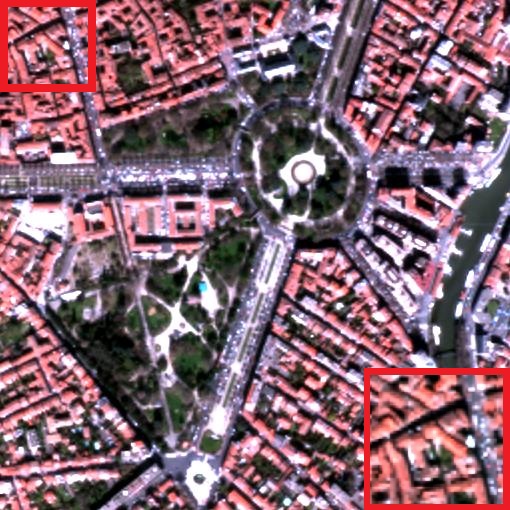}\label{<figure1>}}\mbox{}\vspace{-0.6\baselineskip}
\subfloat{\includegraphics[width=0.21\textwidth]{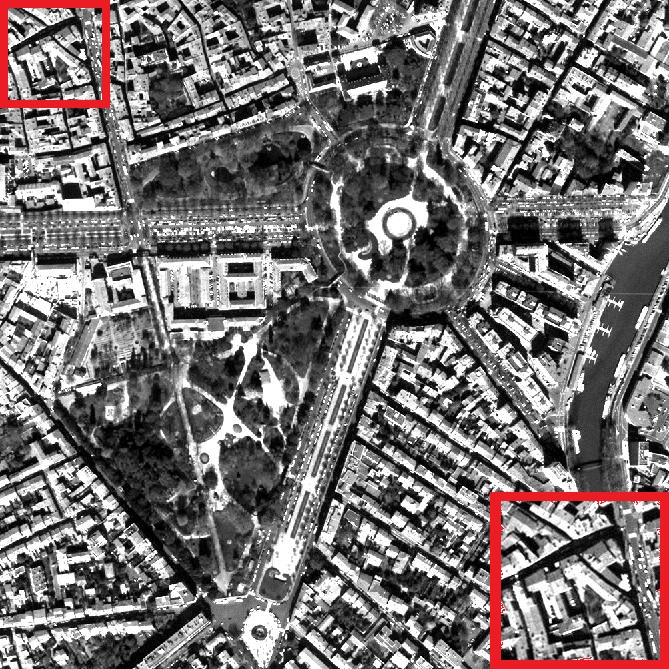}\label{<figure1>}}\mbox{}
\subfloat{\includegraphics[width=0.21\textwidth]{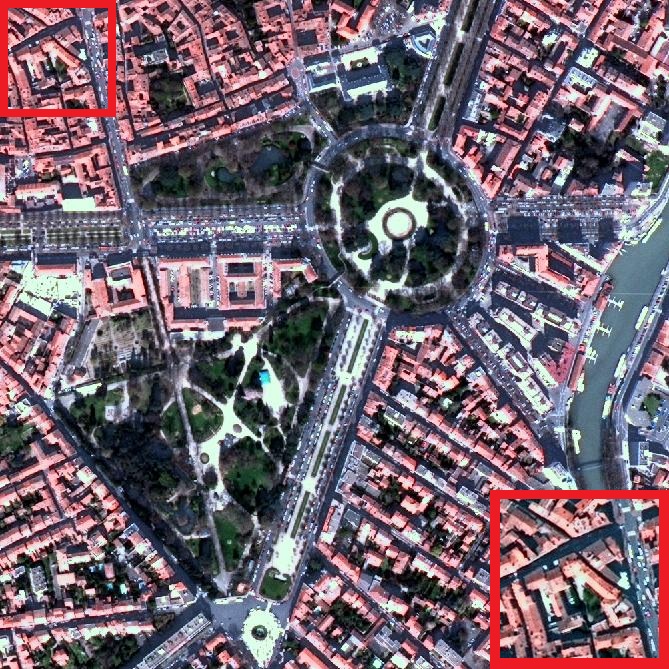}\label{<figure1>}}\mbox{}
\subfloat{\includegraphics[width=0.21\textwidth]{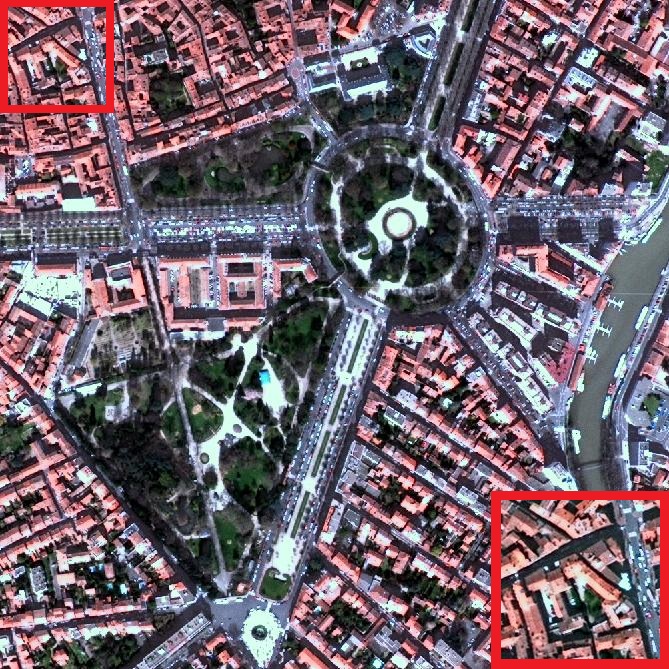}\label{<figure1>}}\mbox{}
\subfloat{\includegraphics[width=0.21\textwidth]{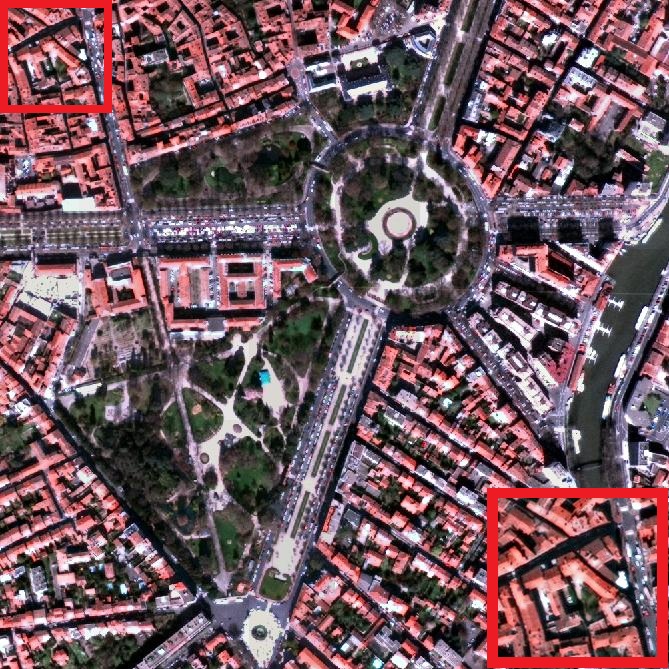}\label{<figure1>}}\mbox{}
\subfloat{\includegraphics[width=0.21\textwidth]{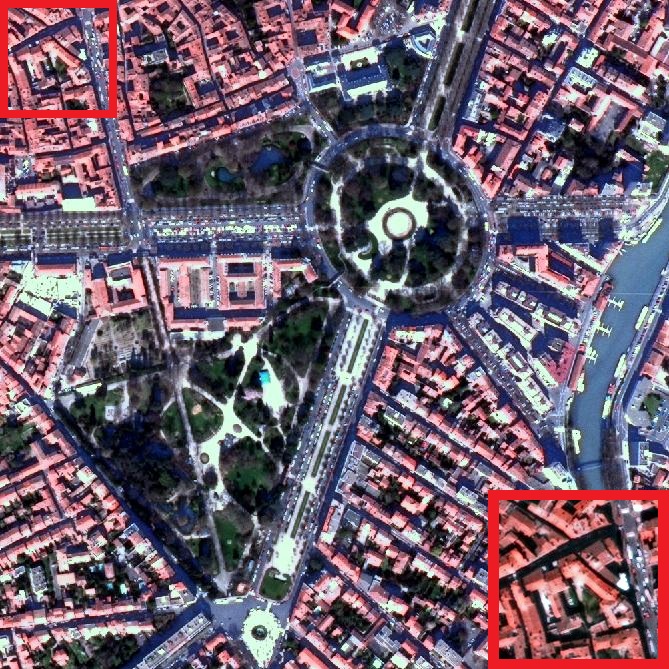}\label{<figure1>}}\mbox{}
\subfloat{\includegraphics[width=0.21\textwidth]{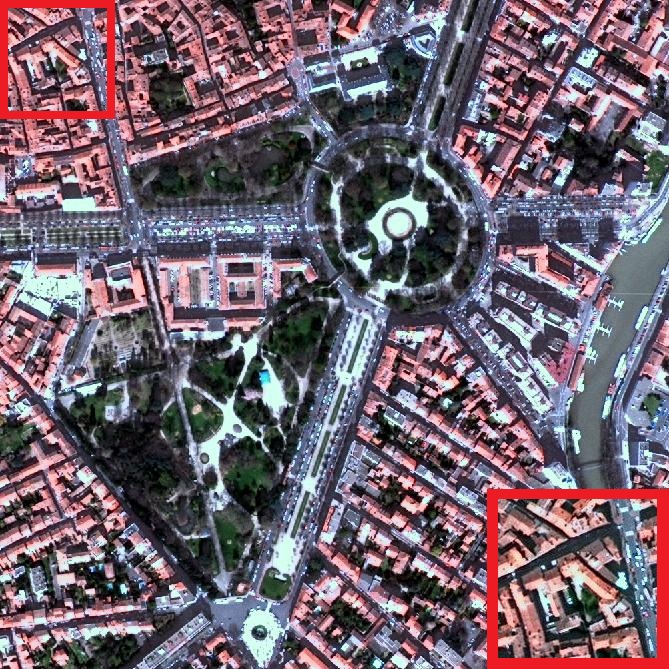}\label{<figure1>}}\mbox{}
\subfloat{\includegraphics[width=0.21\textwidth]{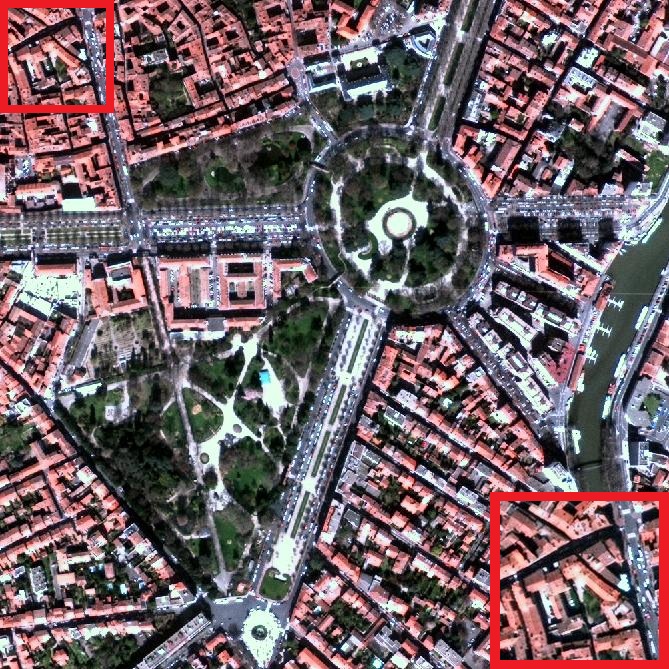}\label{<figure1>}}\mbox{}
\subfloat{\includegraphics[width=0.21\textwidth]{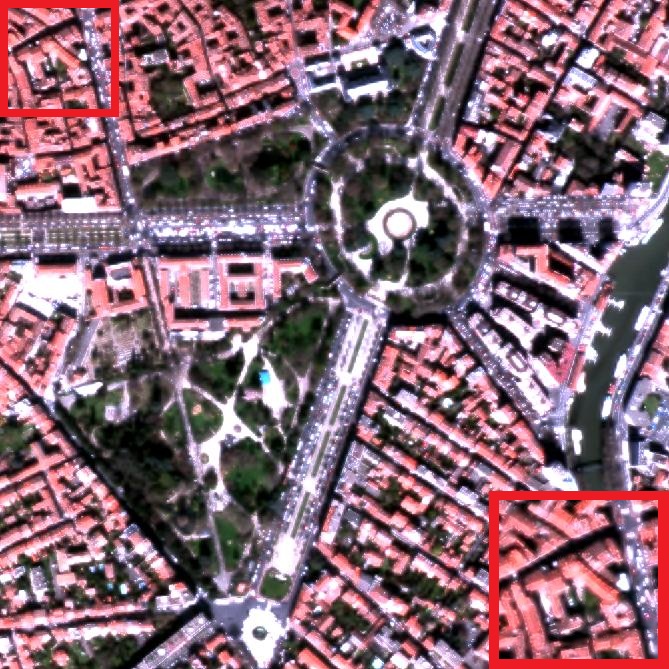}\label{<figure1>}}\mbox{}
\subfloat{\includegraphics[width=0.21\textwidth]{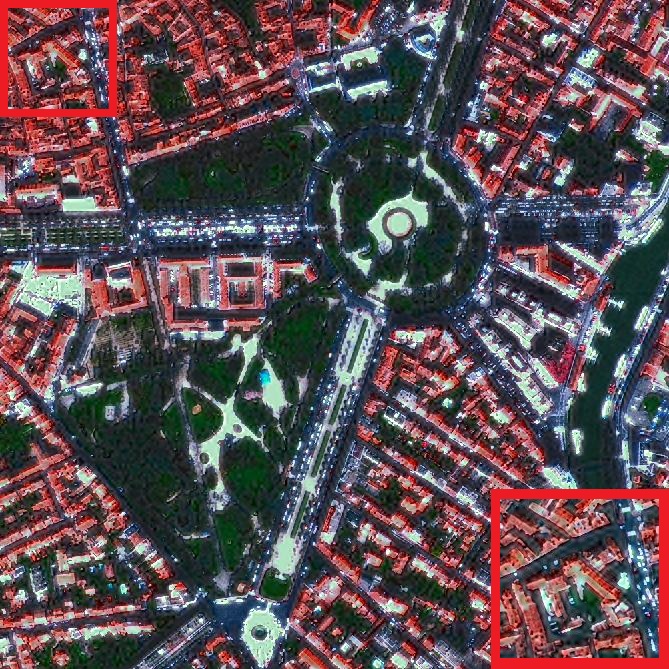}\label{<figure1>}}\mbox{}
\subfloat{\includegraphics[width=0.21\textwidth]{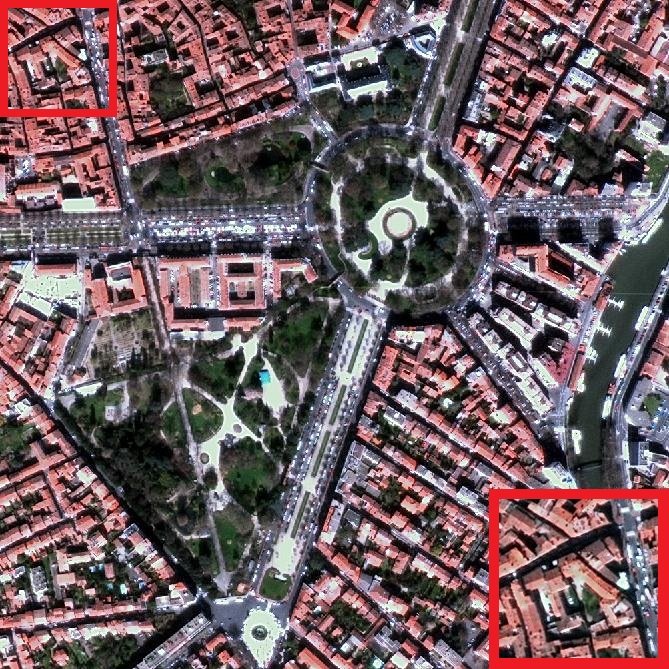}\label{<figure1>}}\mbox{}
\subfloat{\includegraphics[width=0.21\textwidth]{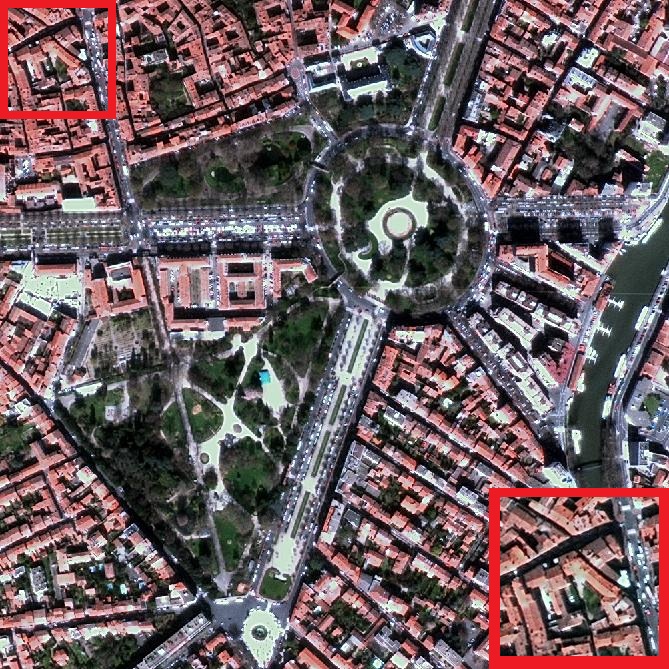}\label{<figure1>}}\mbox{}
\subfloat{\includegraphics[width=0.21\textwidth]{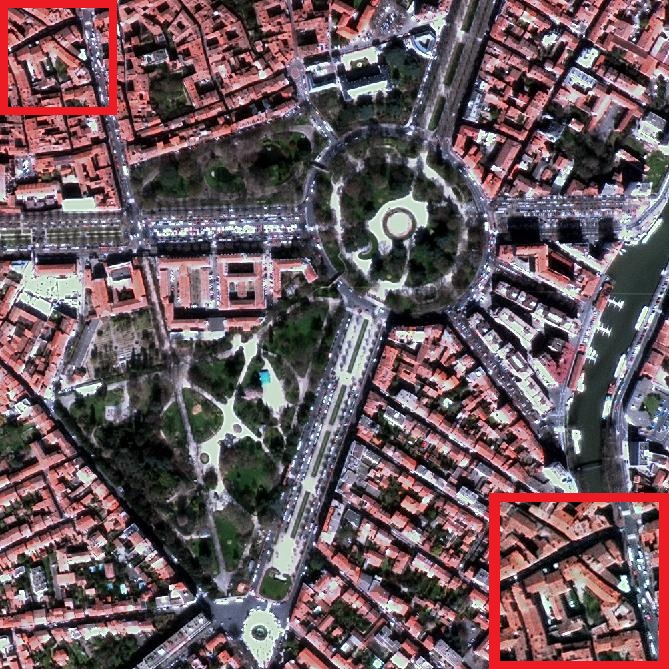}\label{<figure1>}}\mbox{}
\subfloat{\includegraphics[width=0.21\textwidth]{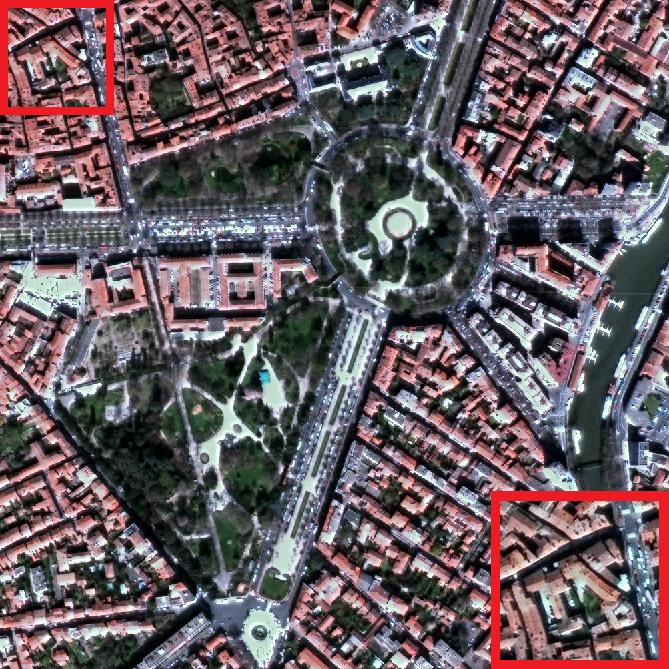}\label{<figure1>}}\mbox{}
\subfloat{\includegraphics[width=0.21\textwidth]{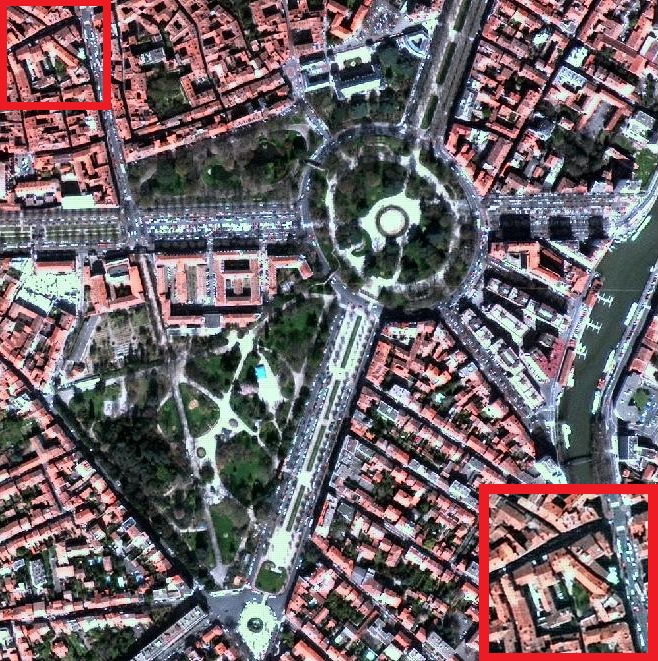}\label{<figure1>}}\mbox{}
\subfloat{\includegraphics[width=0.21\textwidth]{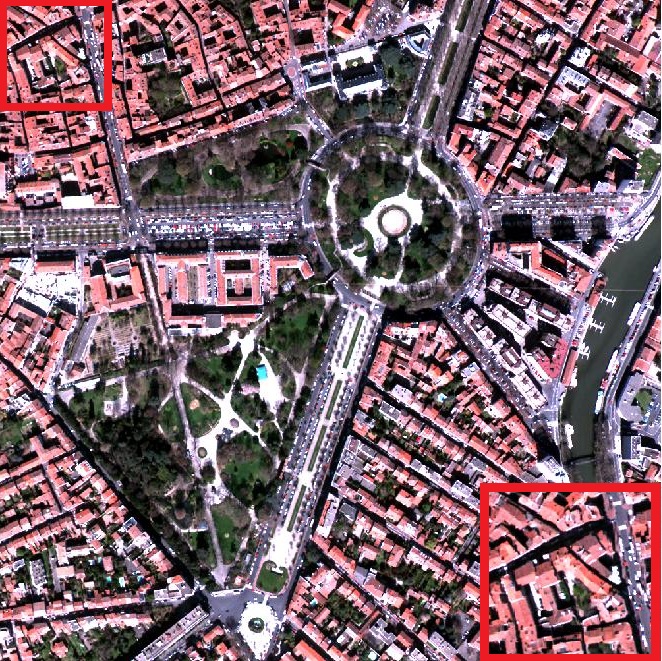}\label{<figure1>}}\mbox{}
\subfloat{\includegraphics[width=0.21\textwidth]{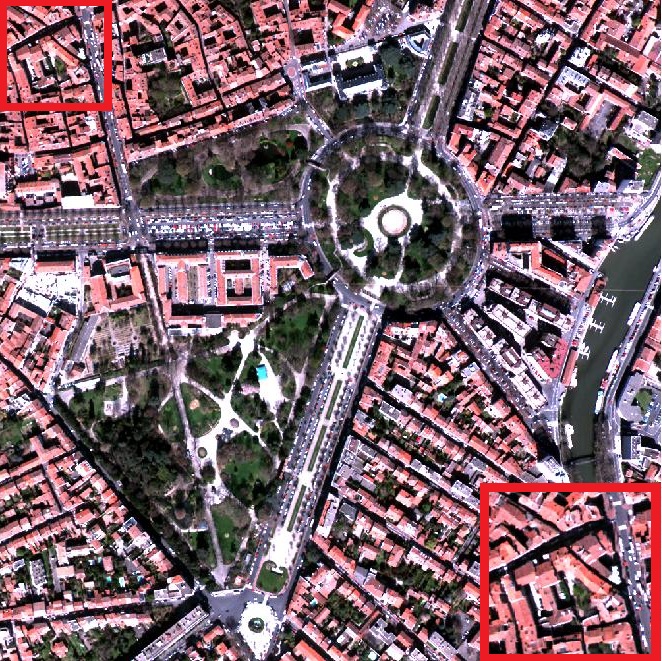}\label{<figure1>}}\mbox{}
\subfloat{\includegraphics[width=0.21\textwidth]{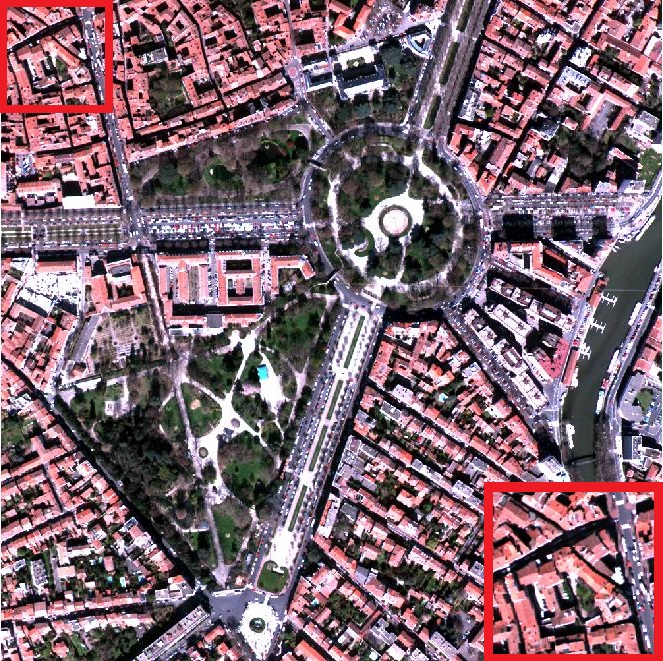}\label{<figure1>}}\mbox{}
\caption{Visual results for Toulouse-Pl\'{e}iades dataset. From top-left to bottom-right: LrMS, PAN, IHS, PCA, BDSD, BT, GS, PRACS, HPF, Indusion, MTF-GLP, MTF-GLP-CBD, MTF-GLP-HPM, MTF-GLP-HPM-PP, CAE, TFNet, DATS, and Reference.}
\label{fig5}
\end{figure*}

% PCA 1989
% IHS 2001
% GLP 2006
% CBD 2007*
% BDSD 2007
% IND 2008*
% HPM-PP 2009
% PRACS 2010
% HPM 2013
% CAE 2019
% TFNET 2020
% DATS

\begin{figure*}[!htb]
\centering
\subfloat{\includegraphics[width=0.21\textwidth]{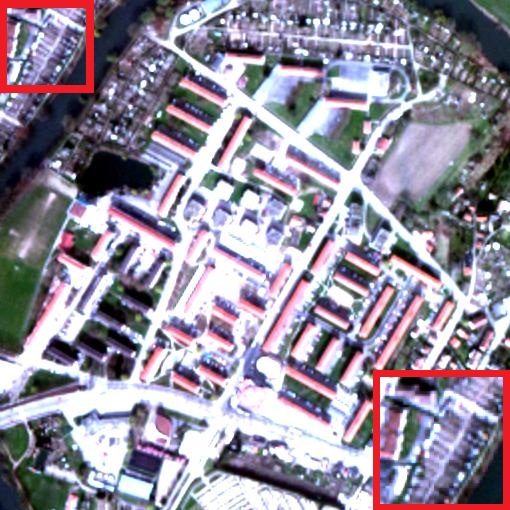}\label{<figure1>}}\mbox{}\vspace{-0.6\baselineskip}
\subfloat{\includegraphics[width=0.21\textwidth]{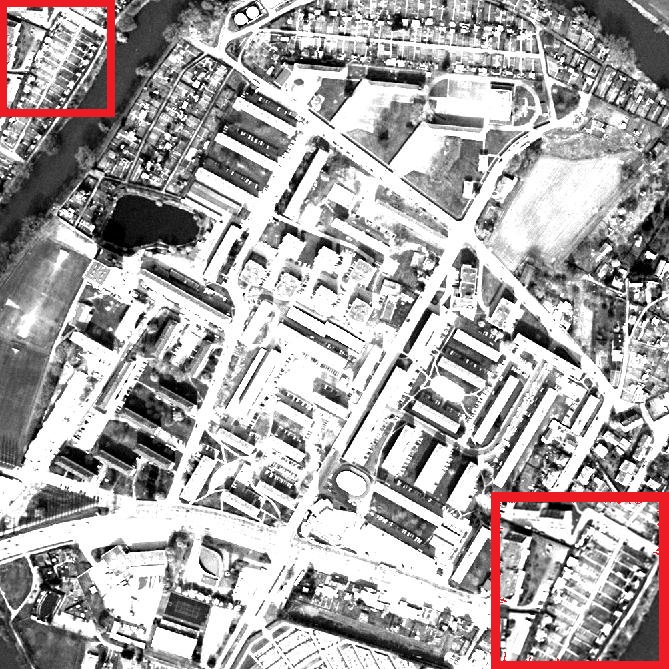}\label{<figure1>}}\mbox{}
\subfloat{\includegraphics[width=0.21\textwidth]{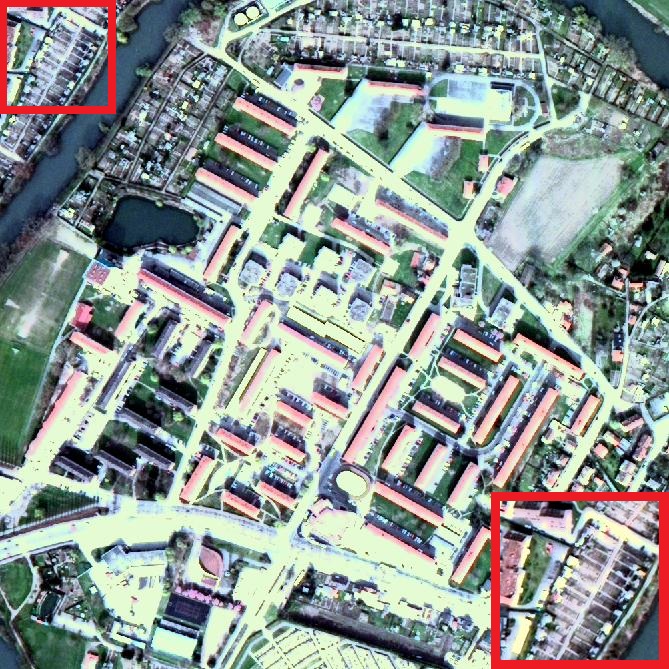}\label{<figure1>}}\mbox{}
\subfloat{\includegraphics[width=0.21\textwidth]{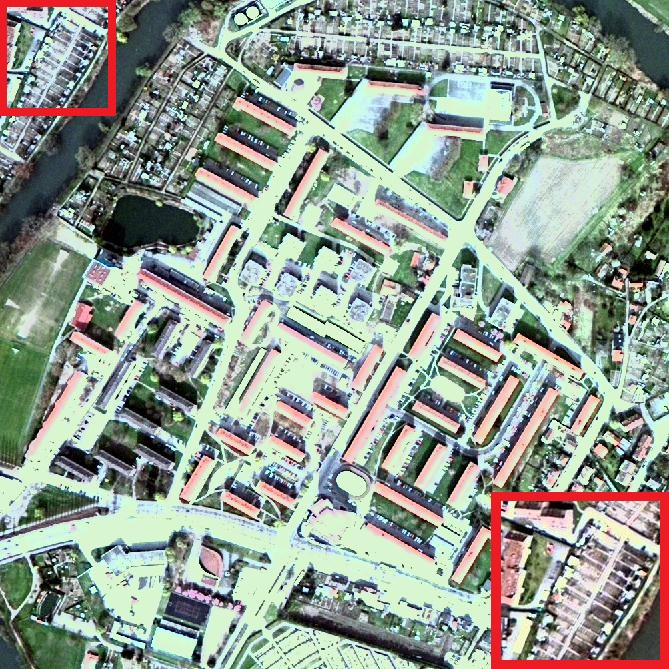}\label{<figure1>}}\mbox{}
\subfloat{\includegraphics[width=0.21\textwidth]{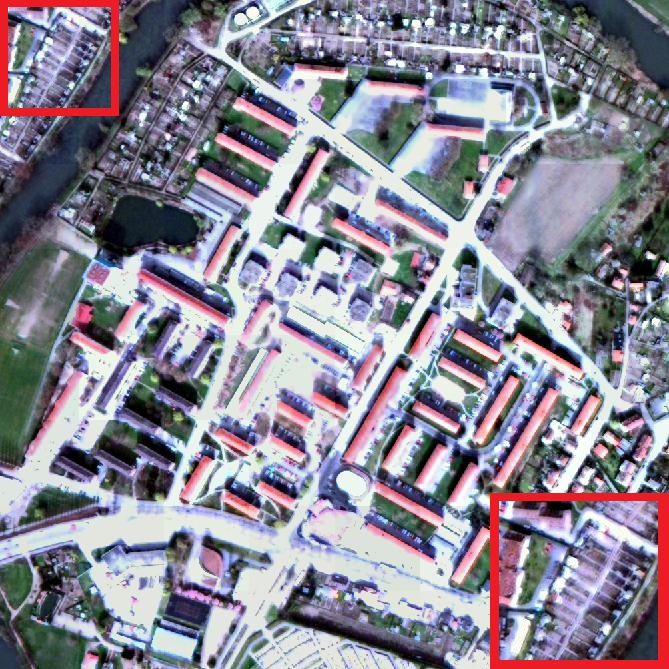}\label{<figure1>}}\mbox{}
\subfloat{\includegraphics[width=0.21\textwidth]{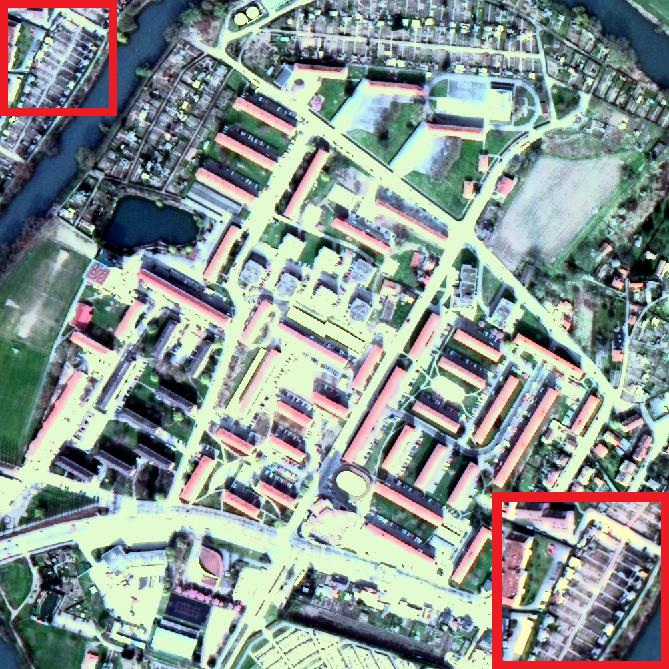}\label{<figure1>}}\mbox{}
\subfloat{\includegraphics[width=0.21\textwidth]{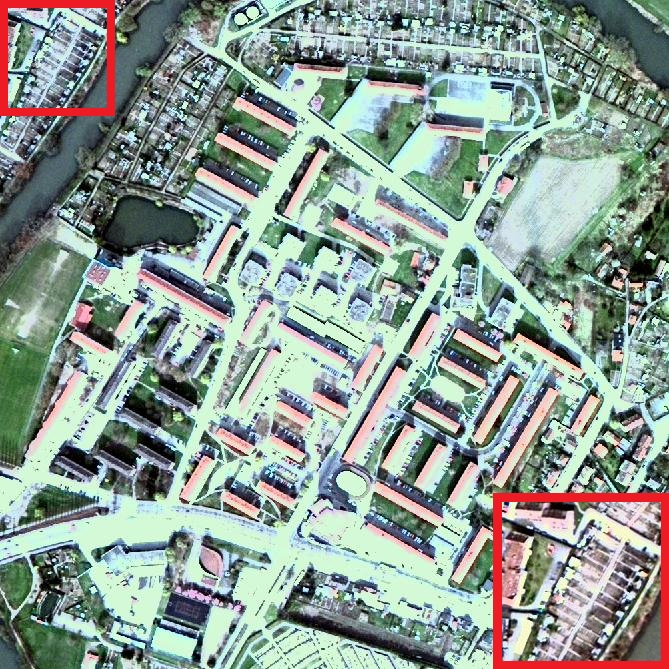}\label{<figure1>}}\mbox{}
\subfloat{\includegraphics[width=0.21\textwidth]{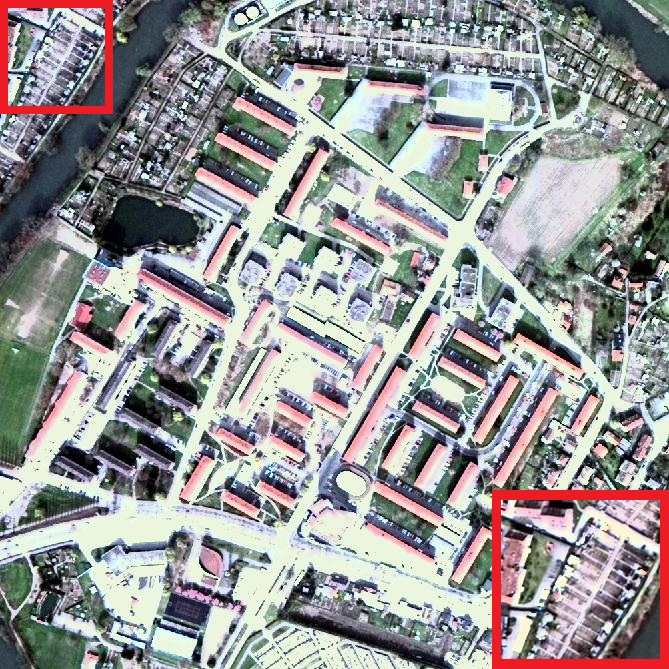}\label{<figure1>}}\mbox{}
\subfloat{\includegraphics[width=0.21\textwidth]{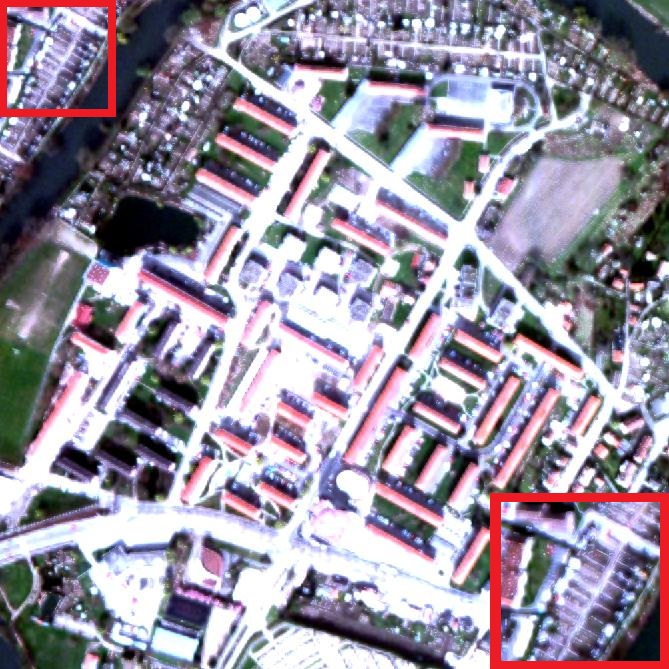}\label{<figure1>}}\mbox{}
\subfloat{\includegraphics[width=0.21\textwidth]{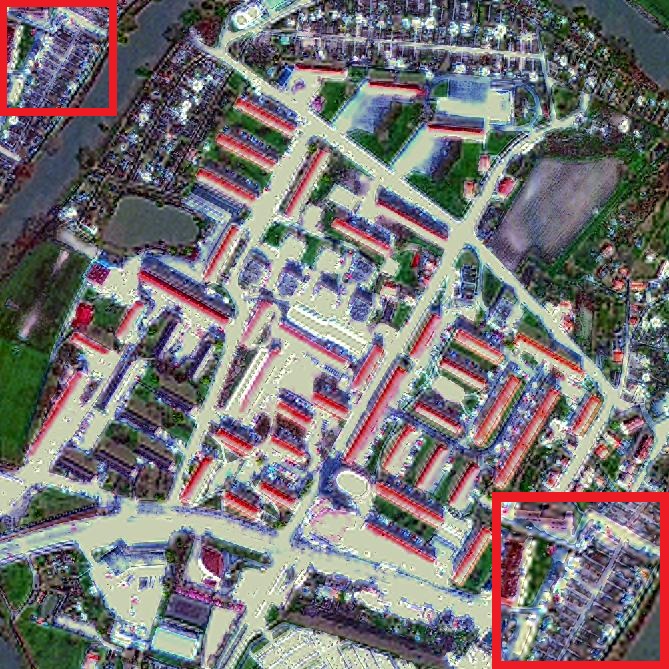}\label{<figure1>}}\mbox{}
\subfloat{\includegraphics[width=0.21\textwidth]{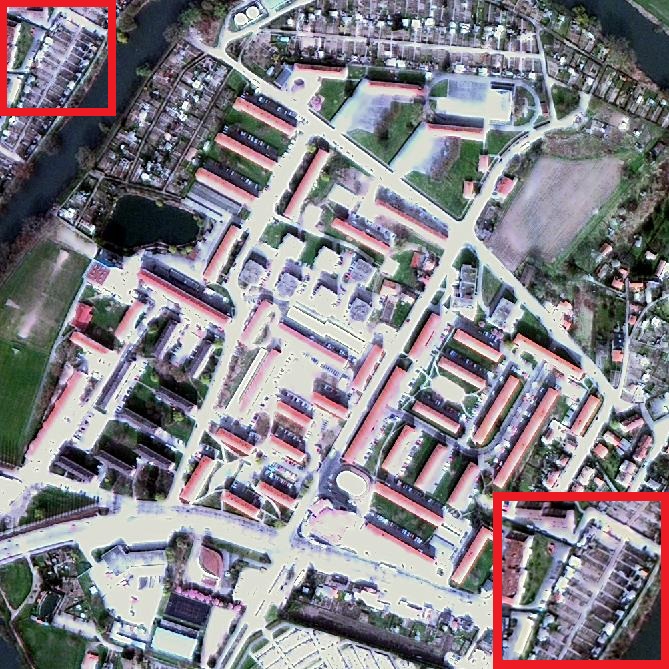}\label{<figure1>}}\mbox{}
\subfloat{\includegraphics[width=0.21\textwidth]{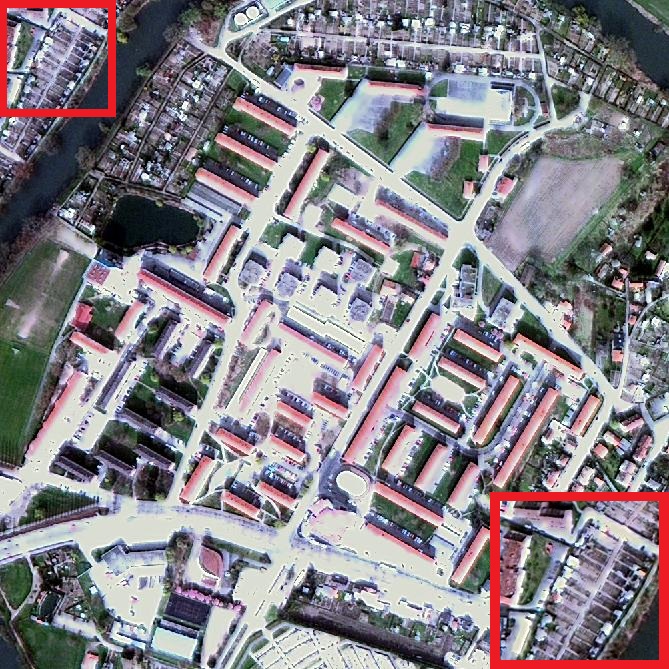}\label{<figure1>}}\mbox{}
\subfloat{\includegraphics[width=0.21\textwidth]{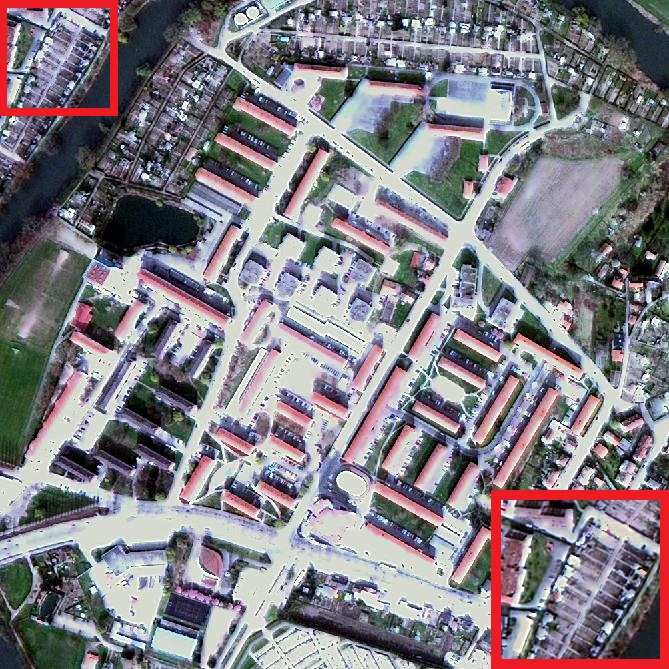}\label{<figure1>}}\mbox{}
\subfloat{\includegraphics[width=0.21\textwidth]{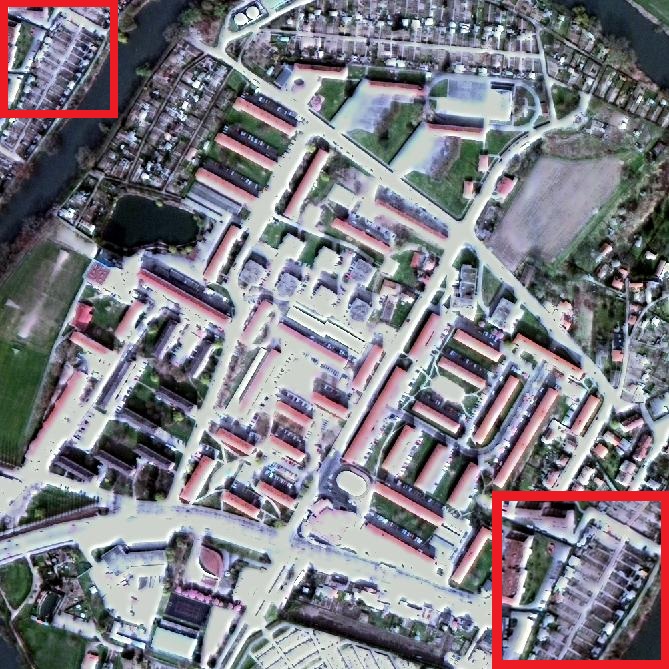}\label{<figure1>}}\mbox{}
\subfloat{\includegraphics[width=0.21\textwidth]{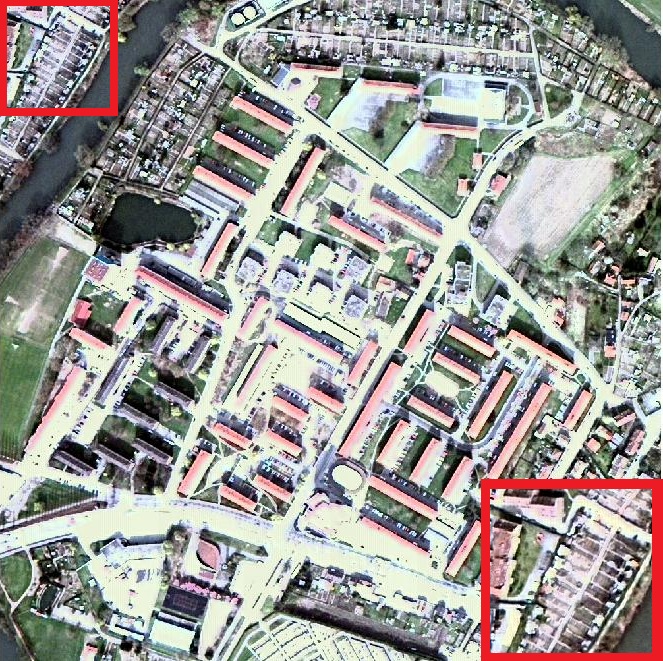}\label{<figure1>}}\mbox{}
\subfloat{\includegraphics[width=0.21\textwidth]{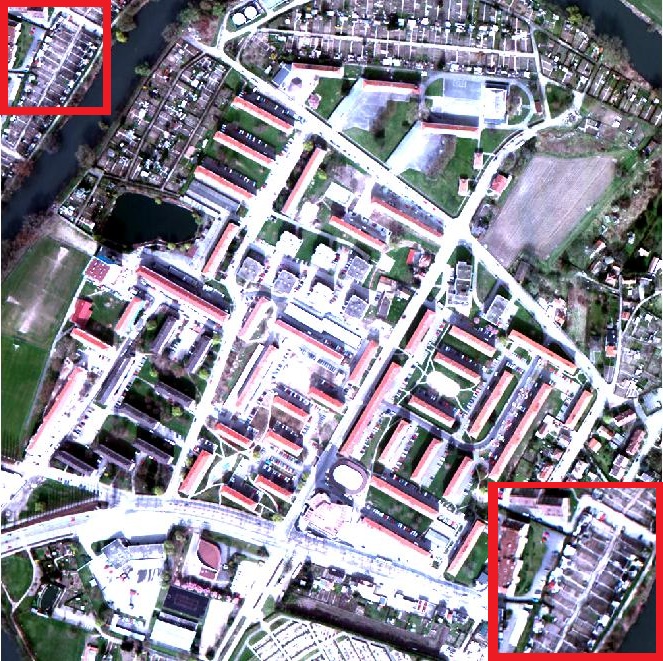}\label{<figure1>}}\mbox{}
\subfloat{\includegraphics[width=0.21\textwidth]{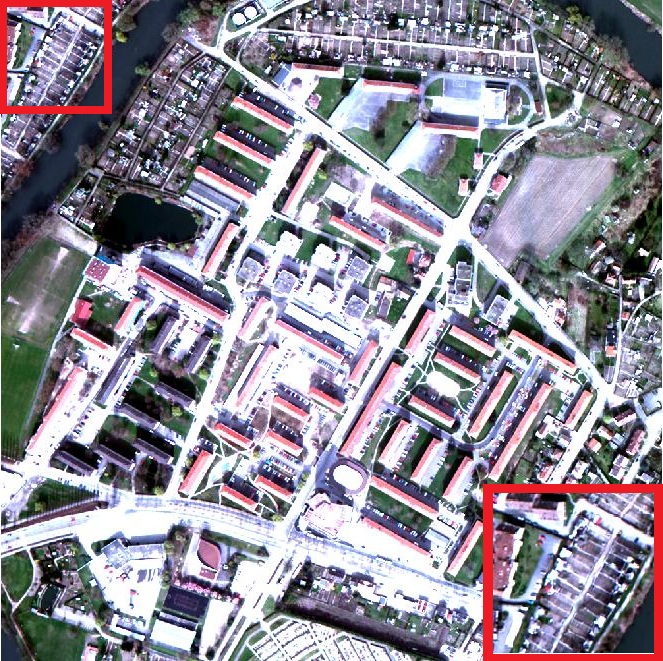}\label{<figure1>}}\mbox{}
\subfloat{\includegraphics[width=0.21\textwidth]{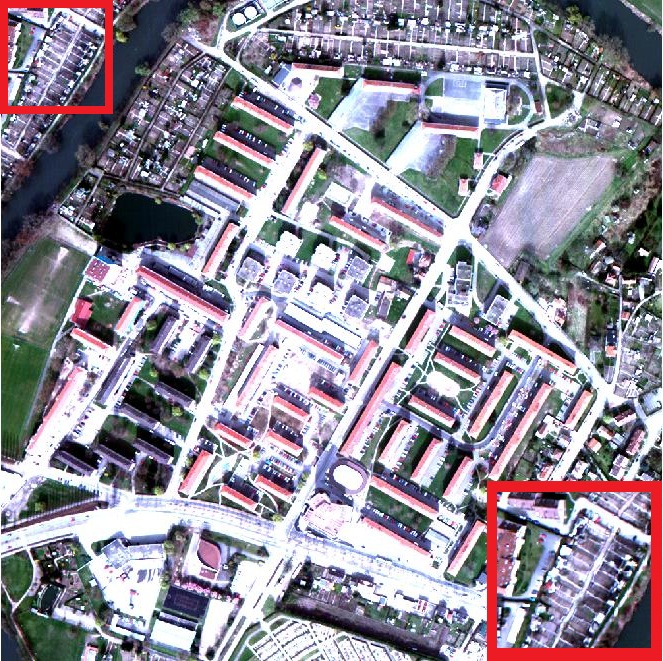}\label{<figure1>}}\mbox{}
\caption{Visual results for Strasbourg-Pl\'{e}iades dataset. From top-left to bottom-right: LrMS, PAN, IHS, PCA, BDSD, BT, GS, PRACS, HPF, Indusion, MTF-GLP, MTF-GLP-CBD, MTF-GLP-HPM, MTF-GLP-HPM-PP, CAE, TFNet, DATS, and Reference.}
\label{fig6}
\end{figure*}

% CAE is also better than most traditional approaches. 
% HPF result lacks enhancing the spatial details in the fused image evident from its SCC metric value.

\begin{figure*}[!htb]
\centering
\subfloat[LrMS]{\includegraphics[width=0.19\textwidth]{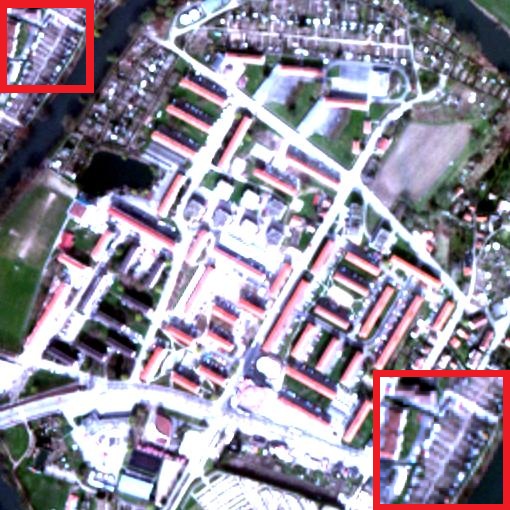}\label{fig7a}}\mbox{}
\subfloat[PAN]{\includegraphics[width=0.19\textwidth]{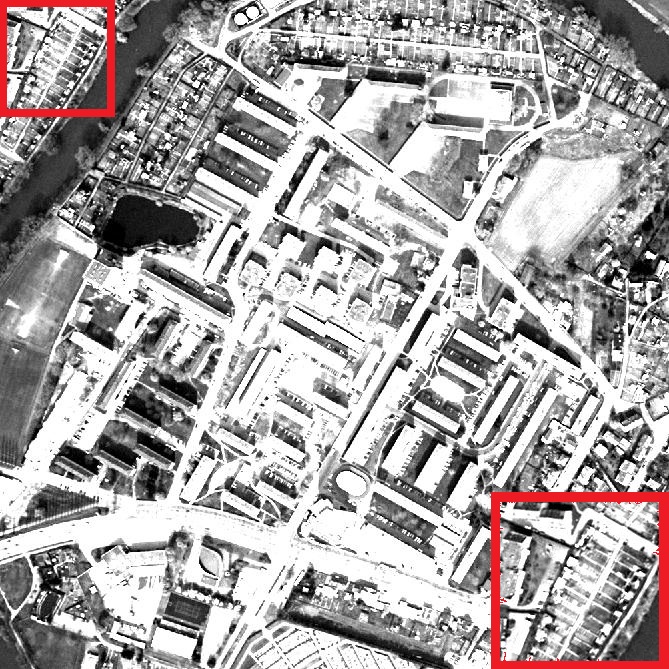}\label{fig7b}}\mbox{}
\subfloat[TFNet]{\includegraphics[width=0.19\textwidth]{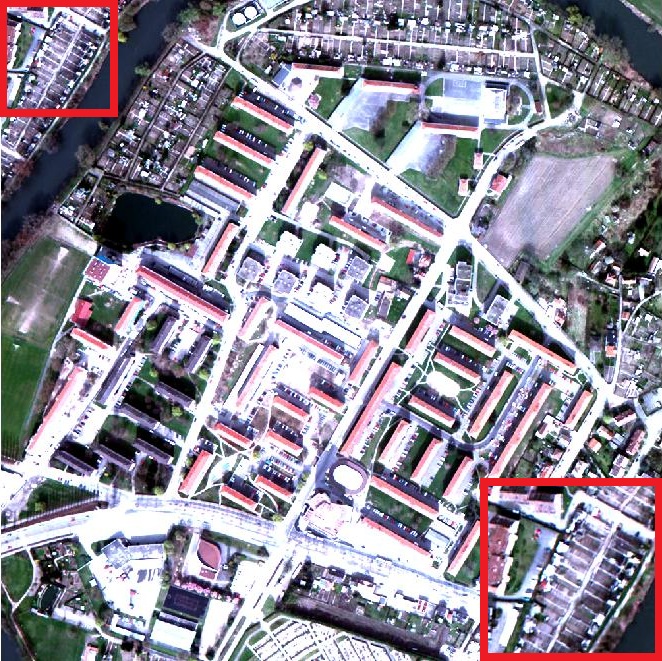}\label{fig7c}}\mbox{}
\subfloat[DATS]{\includegraphics[width=0.19\textwidth]{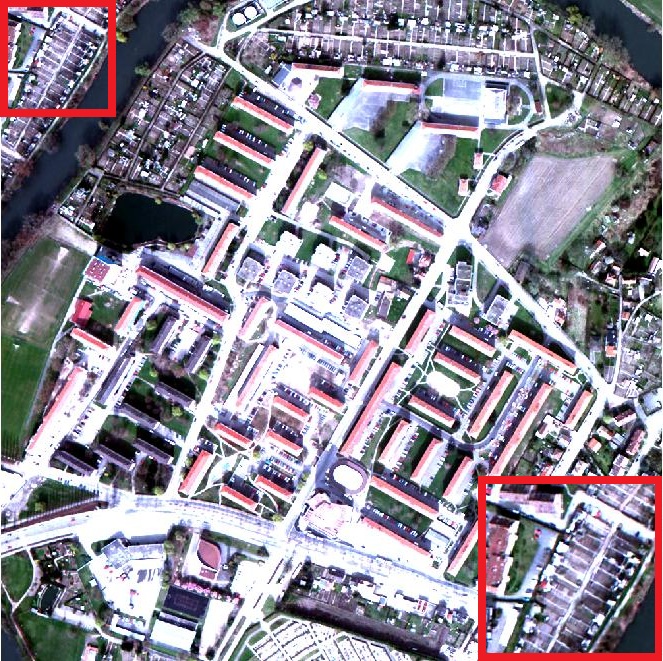}\label{fig7d}}\mbox{}
\subfloat[Reference]{\includegraphics[width=0.19\textwidth]{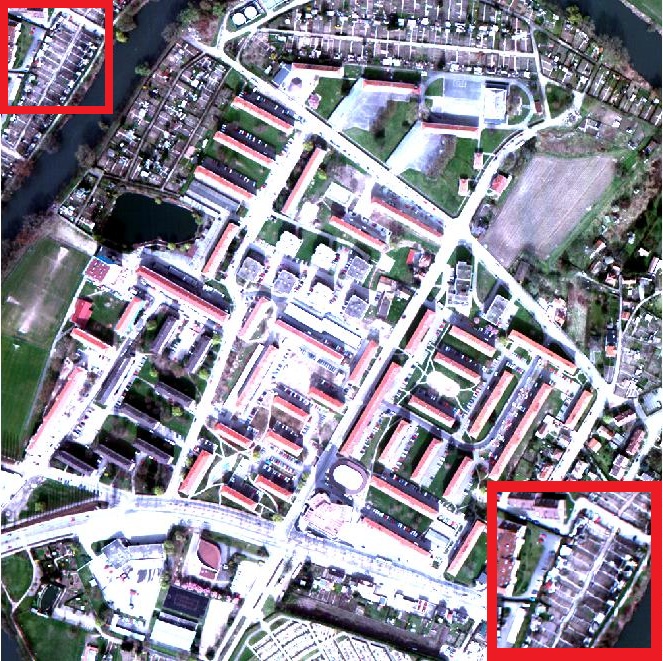}\label{fig7e}}\mbox{}
\caption{Visual results of typical case on Pl\'{e}iades dataset.}
\label{fig7}
\end{figure*}

Fig.~\ref{fig5} and~\ref{fig6} present visual results of different pansharpening algorithms along with the proposed approach in false color synthesis (3 bands as 3, 2, 1). From visual analysis, we observe that the most recent DL-based methods TFNet and DATS are robust in learning sensor characteristics and preserving underlying spectral details while enhancing desired spatial details. They also depict efficiency in terms of image registration and feature learning. However, IHS produced a pansharped image with suitable spatial fidelity but with much spectral distortion. Again the resultant image of indusion suffers from the most distortion whereas BDSD lacks spatial details. Moreover, the visual output of HPM is better compared to its variant HPM-PP. Other approaches including PRACS and GLP achieve promising results among the conventional approaches.

% PCA 1989
% IHS 2001
% GLP 2006
% CBD 2007*
% BDSD 2007
% IND 2008*
% HPM-PP 2009
% PRACS 2010
% HPM 2013
% CAE 2019
% TFNET 2020
% DATS

% In Fig.~\ref{fig5} and~\ref{fig6}, (a) is the upsampled low resolution multispectral image (LrMS) and (b) is its corresponding high resolution panchromatic image (PAN). The remaining (c)-(q) is the pansharpening results produced by different models. Lastly, (r) is the high-resolution multispectral image also known as the reference image.

% HPF results suffer from blurriness. 

In the second experiment, we consider a typical case of evaluation where training and test images are taken by the same sensor but of different locations. We use images of the Pl\'{e}iades dataset representing urban areas of Toulouse and Strasbourg (France), the first image set for training while the second for testing. Notably, this evaluation helps determine the effectiveness of the DL models to learn sensor degradation. Table~\ref{tab4} presents the quality metrics computed for TFNet and the proposed approach along with ideal values for a typical case. The low ERGAS and SAM values for the proposed approach indicate less distortion and high spectral correlation. While the high values of UIQI, SCC and SSIM in the case of the proposed approach shows the model's proficiency learning features for the respective sensor. This results in pansharpened images that are spatially correlated to the reference image. 

% The proposed approach has clearly achieved success in learning the sensor's underlying details and using them for pansharpening corresponding.

% Optimal values obtained are presented in bold.

\begin{table}[!ht]%
\footnotesize
\centering
\caption{List of computed quality metrics for a typical case along with their ideal values. \label{tab4}}
\begin{tabular*}{1.0\linewidth}{@{\extracolsep\fill}lccccccccccclD{.}{.}{3}l@{\extracolsep\fill}}
\toprule
Methods & ERGAS & SAM & UIQI & SCC & SSIM \\
\midrule
Reference   & \textbf{0}    & \textbf{0}   & \textbf{1}    & \textbf{1}   & \textbf{1} \\ 
TFNet       & 0.735        & 0.025       & 0.999        & 0.588      & 0.980 \\ 
DATS        & \textbf{0.735} & \textbf{0.025} & \textbf{0.999} & \textbf{0.593} & \textbf{0.980} \\
\bottomrule
\end{tabular*}
\end{table}

Fig.~\ref{fig7} presents visual results for TFNet and DATS in false color synthesis (3 bands as 3, 2, 1). From visual inspection, we observe the efficacy of the two DL-based methods to extract representative features, to learn fundamental characteristics of sensors, and to preserve underlying spectral fidelity while enhancing required spatial information without manual intervention.

% Fig.~\ref{fig7a} and~\ref{fig7b} are the upsampled low resolution multispectral image (LrMS) and its corresponding high resolution panchromatic image (PAN). Fig.~\ref{fig7c} and~\ref{fig7d} are the pansharpening results produced by TFNet and proposed approach DATS. Lastly, Fig.~\ref{fig7e} is the high-resolution multispectral image as known as the reference image.

% Overall, the visual inspection and quantitative analysis confirm the effectiveness and proficiency of the deep learning-based approaches specifically where separate networks are employed for feature extraction. 

% Overall, they are robust in maintaining the underlying spectral details while enhancing spatial information. These approaches are efficient in learning spatial and spectral features and precisely representing them for generating the optimal pansharpened result. Further addition of attention mechanism ensures the enhanced performance in learning sensor characteristics and significant improvement in results.

\section{Conclusions}
\label{sec:conclusions}

We present a novel dual attention-based two-stream (DATS) pansharpening method that fuses an HR PAN and LrMS image and delivers a pansharped MS image high in both spatial and spectral information. The computed quality measures confirm that the proposed approach can efficiently learn the degradation process of the corresponding sensor and can effectively produce a pansharped image having high spatial details while maintaining the spectral fidelity of the underlying image. In the future, we will explore the transability of our proposed network among different sensors. 

\section*{Acknowledgments}

The authors extend their appreciation to Muhammad Murtaza Khan for helpful suggestions regarding MS/PAN image problem, dataset and comparison.

\bibliographystyle{IEEEtran}
%\addcontentsline{toc}{chapter}{References}
\bibliography{refs}

% {\footnotesize\bibliography{refs}}% common bib file
%% if required, the content of .bbl file can be included here once bbl is generated
%%\input sn-article.bbl

%% Default %%
%%\input sn-sample-bib.tex%

% that's all folks
\end{document}